%% file: ms.tex
\documentclass[letterpaper, 10 pt, journal, twoside]{IEEEtran} 

\IEEEoverridecommandlockouts                              

\usepackage{booktabs} 
\usepackage{amsmath,amsthm,amssymb}
\usepackage{enumerate}
\usepackage{mathtools}
\usepackage[normalem]{ulem} 
\usepackage{graphicx}
\usepackage{color}
\usepackage{units}
\usepackage{comment}
\usepackage{xfrac}
\usepackage{algorithmic}
\usepackage{algorithm}
\usepackage{hyperref}
\usepackage{times}
\input{Notation.tex}

\newcommand{\ReviewEdit}[1]{\textcolor{blue}{#1}}
\newcommand{\ReviewEditRm}[1]{\textcolor{blue}{#1}}
\newcommand{\ReviewRm}[1]{\textcolor{red}{\sout{#1}}}
\newcommand{\ReviewEditBlock}[1]{\color{blue} #1 \color{black}}
\newcommand{\ReviewRmBlock}[1]{\color{red} #1 \color{black}}

\newcommand{\sos}{\textit{SoS}}
\newcommand{\ParHeader}[1]{\textbf{#1}\\}

\newcommand{\Ac}{{\mathcal{A}_c}}

\newcommand{\Ad}{{\mathcal{A}_d}}

\renewcommand{\ParHeader}[1]{}
\renewcommand{\ReviewEdit}[1]{#1}
\renewcommand{\ReviewEditBlock}[1]{#1}
\renewcommand{\ReviewEditRm}[1]{}
\renewcommand{\ReviewRm}[1]{}
\renewcommand{\ReviewRmBlock}[1]{}

\author{Nils Smit-Anseeuw$^{1}$, C. David Remy$^{2}$, and Ram Vasudevan$^{1}$%
\thanks{Manuscript received: February, 24, 2019; Revised June, 4, 2019; Accepted July, 8, 2019.}
\thanks{This paper was recommended for publication by Editor Nikos Tsagarakis upon evaluation of the Associate Editor and Reviewers' comments. 
This work was supported by the National Science Foundation under Grant No. 1562612. Any opinion, findings, and conclusions or recommendations expressed in this material are those of the authors and do not necessarily reflect the views of the National Science Foundation.} 
\thanks{$^{1}$Nils Smit-Anseeuw and Ram Vasudevan are with the Department of Mechanical Engineering at the University of Michigan, Ann Arbor
        {\tt\footnotesize nilssmit@umich.edu, ramv@umich.edu}}%
\thanks{$^{2} $C. David Remy is with the Institute for Nonlinear Mechanicas at the University of Stuttgart
        {\tt\footnotesize david.remy@inm.uni-stuttgart.de}}%
\thanks{Digital Object Identifier (DOI): see top of this page.}
}

\title{Walking with Confidence: \\Safety Regulation for Full Order Biped Models}

\begin{document}
\maketitle

\begin{abstract}
Safety guarantees are valuable in the control of walking robots, as falling can be both dangerous and costly.
Unfortunately, set-based tools for generating safety guarantees (such as sums-of-squares optimization) are typically restricted to simplified, low-dimensional models of walking robots.
For more complex models, methods based on hybrid zero dynamics can ensure the local stability of a pre-specified limit cycle, but provide limited guarantees.  
This paper combines the benefits of both approaches by using sums-of-squares optimization on a hybrid zero dynamics manifold to generate a guaranteed safe set for a 10-dimensional walking robot model.
Along with this set, this paper describes how to generate a controller that maintains safety by modifying the manifold parameters when on the edge of the safe set.
The proposed approach, which is applied to a bipedal Rabbit model, provides a roadmap for applying sums-of-squares \ReviewRm{verification} techniques to high dimensional systems.
This opens the door for a broad set of tools that can generate \ReviewRm{safety guarantees and regulating}\ReviewEdit{flexible and safe} controllers for complex walking robot models.

\end{abstract}


  \input{sections/Introduction}
  
  \input{sections/Methods}

  \input{sections/Results}

  \input{sections/Discussion}

\bibliographystyle{ieeetr}
\bibliography{cdc2017}

\end{document}

%% file: Notation.tex
\newcommand{\q}{\ensuremath{q}}
\newcommand{\dq}{    \ensuremath{\dot{q}}}
\newcommand{\xF}{     \ensuremath{x}}
\newcommand{\dxF}{     \ensuremath{\dot{x}}}
\newcommand{\inptF}{  \ensuremath{u}}
\newcommand{\fF}{     \ensuremath{f}}
\newcommand{\gF}{     \ensuremath{g}}
\newcommand{\rxF}{    \ensuremath{\Delta}}
\newcommand{\rqF}{    \ensuremath{\Delta_q}}
\newcommand{\rdqF}{   \ensuremath{\Delta_{\dot{q}}}}
\newcommand{\sF}{     \ensuremath{S}}
\newcommand{\qspaceF}{\ensuremath{Q}}
\newcommand{\qfeas}{\ensuremath{\tilde{Q}}}
\newcommand{\dqfeas}{\ensuremath{T\tilde{Q}}}

\newcommand{\dqspaceF}{\ensuremath{TQ}}

\newcommand{\uspaceF}{\ensuremath{U}}

\newcommand{\nqF}{    \ensuremath{n_q}}
\newcommand{\nuF}{    \ensuremath{n_u}}
\newcommand{\inptMaxF}{\ensuremath{u_{max}}}
\newcommand{\dqMax}{\ensuremath{\dot{q}_{max}}}

\newcommand{\naS}{    \ensuremath{n_{\alpha}}}
\newcommand{\aS}{     \ensuremath{\alpha}}
\newcommand{\daS}{    \ensuremath{\dot{\alpha}}}
\newcommand{\xS}{     \ensuremath{x_\alpha}}
\newcommand{\dxS}{     \ensuremath{\dot{x}_\alpha}}
\newcommand{\inptS}{  \ensuremath{u_\alpha}}
\newcommand{\fS}{     \ensuremath{f_\alpha}}
\newcommand{\gS}{     \ensuremath{g_\alpha}}

\newcommand{\aspaceS}{\ensuremath{A}}

\newcommand{\daspaceS}{\ensuremath{TA}}
\newcommand{\uspaceS}{\ensuremath{U_\alpha}}

\newcommand{\xHZD}{\ensuremath{\hat{x}}}
\newcommand{\nHZD}{\ensuremath{n_z}}
\newcommand{\dxHZD}{\ensuremath{\dot{\hat{x}}}}
\newcommand{\inptHZD}{\ensuremath{\hat{u}}}
\newcommand{\qfHZD}{\ensuremath{q_0}}
\newcommand{\dqfHZD}{\ensuremath{\dot{q}_0}}
\newcommand{\thHZD}{\ensuremath{\theta}}
\newcommand{\dthHZD}{\ensuremath{\dot{\theta}}}
\newcommand{\targHZD}{\ensuremath{h}}
\newcommand{\diffHZD}{\ensuremath{\Phi}}
\newcommand{\zspaceHZD}{\ensuremath{Z}}

\newcommand{\fHZD}{\ensuremath{\hat{f}}}
\newcommand{\gHZD}{\ensuremath{\hat{g}}}
\newcommand{\rHZD}{\ensuremath{\hat{\Delta}}}
\newcommand{\sHZD}{\ensuremath{\hat{S}}}

\newcommand{\uspaceHZD}{\ensuremath{\hat{U}}}
\newcommand{\fspaceHZD}{\ensuremath{\zspaceHZD^{F}}}

\newcommand{\ctrlF}{  \ensuremath{u_s}}
\newcommand{\vdF}{    \ensuremath{V}}
\newcommand{\vdHZD}{  \ensuremath{\hat{V}}}
\newcommand{\bfHZD}{  \ensuremath{\hat{v}}}
\newcommand{\ctrlHZD}{\ensuremath{\hat{u}_s}}

\newcommand{\fPoly}{\ensuremath{\hat{f}_p}}

\newcommand{\guPoly}{\ensuremath{\hat{g}_{p}}}
\newcommand{\gdPoly}{\ensuremath{\hat{e}_{p}}}
\newcommand{\rPoly}{\ensuremath{R_p}}
\newcommand{\qPoly}{\ensuremath{q}}

\newcommand{\lPoly}{\ensuremath{\lambda}}
\newcommand{\sPoly}[1]{\ensuremath{\sigma_{#1}}}

\newcommand{\rawctrlPoly}{\ensuremath{\hat{u}_0}}
\newcommand{\maskctrlPoly}{\ensuremath{\hat{u}_r}}
\newcommand{\wrawPoly}{\ensuremath{w_0}}
\newcommand{\wsafePoly}{\ensuremath{w_s}}
\newcommand{\vmPoly}{\ensuremath{v_m}}

\newcommand{\real}{\mathbb{R}}
\newcommand{\transpose}{\ensuremath{^\top}}
\newcommand{\lfx}{\mathcal{L}_f^{x}}
\newcommand{\lgx}{\mathcal{L}_g^{x}}
\newcommand{\lfa}{\mathcal{L}_{f_\alpha}^{x_\alpha}}
\newcommand{\lga}{\mathcal{L}_{g_\alpha}^{x_\alpha}}
\newcommand{\lfPoly}{\mathcal{L}_{\hat{f}_p}^{\xHZD}}
\newcommand{\lguPoly}{\mathcal{L}_{\hat{g}_p}^{\xHZD}}
\newcommand{\lgdPoly}{\mathcal{L}_{\hat{e}_p}^{\xHZD}}
\usepackage{accents}

\newcommand{\Ram}[1]{{\normalsize{\textbf{({\color{green}Ram:\ }#1)}}}}

\newcommand{\David}[1]{{\normalsize{\textbf{({\color{red}David:\ }#1)}}}} 
\newcommand{\Nils}[1]{{\normalsize{\textbf{({\color{blue}Nils:\ }#1)}}}}

\newcommand{\Dan}[1]{{\normalsize{\textbf{({\color{magenta}Dan:\ }#1)}}}}

\newtheorem{thm}{Theorem}

\newtheorem{definition}{Definition}

\newtheorem{hhyp}{HH}
\newtheorem{example}{Example}
\newtheorem{constraint}{SoS Constraint}

\newtheorem*{viability}{Viability Conditions}

%% file: sections/Introduction.tex
\section{Introduction}
\ParHeader{Safety is important for walking robots, but challenging.}
\IEEEPARstart{A}{voiding} falls is a safety critical and challenging task for legged robotic systems.
This challenge is compounded by strong limits on the available actuation torques; particularly at the ankle or  ground contact point. 
These limits in actuation mean that the motion of a legged robot is often dominated by its mechanical dynamics, which are hybrid, nonlinear, and unstable.
A consequence of these limitations is that a controller might be required to take a safety preserving action well before the moment a failure occurs.

Consider, for example, a bipedal robot that just entered single stance during a fast walking gait.
The robot is pivoting dynamically over the stance foot and can only apply limited ankle torques to control its motion.  
To catch the robot again, the swing foot needs to be brought forward rapidly and be placed well in front of the robot.
If the forward velocity of the robot and hence the pivoting motion is too fast, there will not be enough time to complete this foot placement far enough in front of the stance leg to slow the robot down \cite{pratt2006capture}.
As a result, the robot's speed increases further, leaving even less time for leg swing in the subsequent steps.
The robot might manage to complete another couple of strides, but at this point a fall is inevitable and no control action can prevent it.

\begin{figure}
    \centering
    \includegraphics[width=\columnwidth]{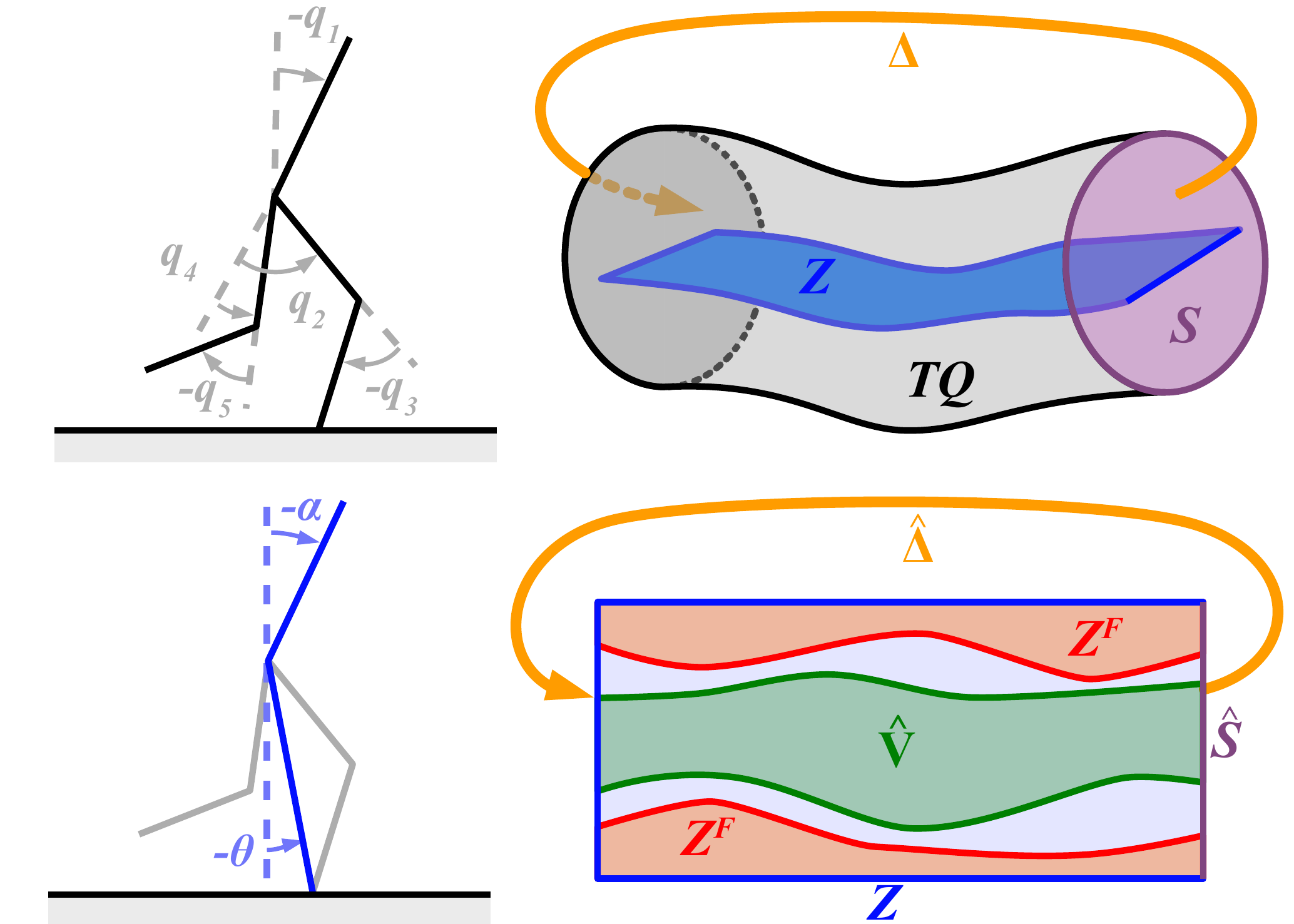}
    \vspace{-0.5cm}
    \caption{Generating safety guarantees for a high dimensional robot (illustrated on Rabbit \cite{chevallereau2003rabbit}).
    The state-space of the full robot is given in the top right figure, where $\dqspaceF$ is the tangent space on $\qspaceF$, $\sF$ is the hybrid guard representing foot touchdown, and $\rxF$ is the corresponding discrete reset map.
    Using feedback linearization, we restrict our states to lie on a low-dimensional manifold $\zspaceHZD$, reducing the state-space dimension to an amenable size for sums-of-squares analysis.
    This manifold is parameterized by \ReviewEdit{the underactuated degrees of freedom of the robot $\thHZD$, as well as a set of shaping parameters $\aS$}\ReviewRm{a set of states that are important for safety (e.g. for Rabbit we chose the pitch angle $\aS$ and stance leg angle $\thHZD$)}.
    The shaping parameters can be modified in real-time by a control input, allowing for \ReviewEdit{a broad range of}\ReviewRm{safety regulating} behaviours on $\zspaceHZD$.
    To guarantee safety on $\zspaceHZD$ we find the set of unsafe states $\fspaceHZD$ from which the state may leave the manifold (for instance due to motor torque limits).
    We then use sums-of-squares tools \cite{parrilo2000structured} to find a control invariant set $\vdHZD \subset \zspaceHZD \setminus \fspaceHZD$.
    This control invariant set can be used to define a semi-autonomous, guaranteed safe controller for the full robot dynamics.}
    \label{fig:approach}
    \vspace{-0.6cm}
\end{figure}

Knowing the limits of safe operation is akin to knowing the set of states from which falls, even in the distant future, can be avoided.
Such knowledge is valuable for many reasons.
Knowing that a fall is inevitable is useful in itself, as it allows a robot to brace for the imminent impact.
Knowing the distance from the border of the safe set could allow a robot to estimate the set of impulses that can be withstood without failing.
This would allow it to judge whether or not it can safely interact with the environment in a given situation; for example, to push a cart while walking.
\ReviewRm{In terms of control, the value lies in the flexibility this knowledge can create.}
\ReviewEdit{Most importantly, this knowledge is valuable} due to the flexibility it can create.
Rather than stabilizing the robot motion along a specified trajectory, one could imagine controllers that are adaptive to adjust to the environment, to maximize performance, or to fulfill a secondary task such as pointing a sensor onto a target.
Any of these secondary tasks can be pursued as long as the state of the robot is within the safe set.

In this context, a representation of the set of safe states enables the construction of a regulator that monitors the system state and takes safety preserving actions only when the robot is at risk of failure \cite{wieland2007constructive}.
Such a regulator could guarantee safe operation, while allowing a secondary control system to behave flexibly as long as safety is not threatened.
\ParHeader{Sums of squares broad overview, curse of dimensionality}
Identifying such safety limits, however, is a challenging problem for nonlinear and hybrid systems.
A promising tool for identifying the safety limits of a legged robotic system is sums-of-squares (SoS) optimization \cite{parrilo2000structured}.
This approach uses semi-definite programming to identify the limits of safety in the state space of a system as well as associated controllers for a broad class of nonlinear \cite{majumdar2014,henrion2014,korda2016} and hybrid systems \cite{prajna,shia2014convex}.
These safe sets can take the form of \emph{reachable sets} (sets that can reach a known safe state) \cite{koolen2016balance,shia2014convex,majumdar2014} or \emph{invariant sets} (sets whose members can be controlled to remain in the set indefinitely) in state space \cite{wieber2002stability,prajna,posa2017balancing}.
However, the representation of each of these sets in state space severely restricts the size of the problem that can be tackled by these approaches.
\ParHeader{How is SoS currently used (hint: simple models)}
To accommodate this limitation, sums-of-squares analysis has been primarily applied to reduced models of walking robots: ranging from spring mass models \cite{zhao2017optimal}, to inverted pendulum models \cite{koolen2016balance,tang2017invariant} and  to inverted pendulum models with an offset torso mass \cite{posa2017balancing}.
The substantial differences between these simple models and real robots causes difficulty when applying these results to hardware.

\ParHeader{Introduction to HZD as dimensionality reducer}
A contrasting approach to designing stable controllers for high dimensional, underactuated robot models uses hybrid zero dynamics (HZD) \cite{westervelt}.
In this approach, feedback linearization is used to drive the actuated degrees of freedom of the robot towards a lower dimensional hybrid zero dynamics manifold.
This manifold is specified as the zero levelset of a configuration-dependent output vector and represents the motion of the robot in its underactuated degrees of freedom.

\ParHeader{Safety of HZD}
Significant progress has been made in the generation of safety certificates for HZD controllers.
Much of this work \cite{ames2014rapidly,hsu2015control,nguyen2015optimal,nguyen2016exponential,nguyen2016dynamic} relies on the Poincar\'e stability of a periodic limit cycle in order to generate safety guarantees.
This reliance is restrictive, as it precludes\ReviewRm{ recovery} behaviors that would leave the neighborhood of the limit cycle.
%
Recent work has been done to extend the range of safe HZD behaviours beyond a single limit cycle neighborhood \cite{motahar2016composing,veer2018safe,ames2017first}.
\ReviewEdit{In \cite{motahar2016composing} and \cite{veer2018safe}, the controller is allowed to discretely switch between a family of periodic gaits.
Safety is then ensured using a dwell time constraint that limits how frequently switching can occur.}
In \cite{ames2017first}, a combination of HZD and finite state abstraction is used to safely regulate forward speed of a \emph{fully-actuated} bipedal robot in continuous time.
Our approach shares similarities with these recent papers, but allows for continuous-time variation of behaviour (instead of discrete switching), and applies to \emph{underactuated} robotic systems.

\ParHeader{Overall approach: HZD plus SoS}
In this paper, we build on both of these broad approaches to safety and control synthesis for legged robotic systems.
To combine the full-model accuracy of hybrid zero dynamics and the set-based safety guarantees of sums-of-squares programming, we propose the following approach (Fig.~\ref{fig:approach}).
First, we use hybrid zero dynamics to map the full order dynamics to a low dimensional hybrid manifold.
We control the dynamics on the manifold using a set of \emph{shaping parameters}, which are modified in continuous time to \ReviewEdit{modify robot behaviour}\ReviewRm{regulate safety}.
We then use sums-of-squares programming to find a subset of this manifold which can be rendered forward control invariant.
Once this subset is found on the low dimensional manifold, a regulator can be constructed that allows for free control of the manifold dynamics when safety is not at risk, but switches to a safety preserving controller when safety is threatened.

The approach is presented in a general form that extends to a large class of underactuated bipedal robots.
Throughout the paper, an example implementation is given for a 10-dimensional model of the robot Rabbit \cite{chevallereau2003rabbit} and a tracking task is used to illustrate semi-autonomous safe control.
To the best of our knowledge, this is the highest dimensional walking robot model for which set-based safety guarantees have been generated thus far.

\ParHeader{Paper Structure}

The rest of this paper is organized as follows:
Section \ref{sec:formulation} formally defines the assumptions and objective of this paper.
The next two sections describe our method.
Section~\ref{sec:HZD} constructs a low dimensional zero dynamics manifold with control input.
In Section~\ref{sec:SOS} we present a sums-of-squares optimization which finds a control invariant subset of the manifold that avoids a designated set of unsafe states.
Section~\ref{sec:results} describes the results of our implementation on the robot Rabbit \cite{chevallereau2003rabbit}, and conclusions are presented in Section~\ref{sec:conclusion}.


%% file: sections/Methods.tex
\section{Problem Setup}
\label{sec:formulation}

\subsection{Robot Model}

For simplicity, we apply similar modeling assumptions to those made in \cite{westervelt}.
\ReviewRm{We begin by assuming robot hypotheses RH1-6 formally stated in \cite{westervelt}.} That is, the robot is modeled as a planar chain of rigid links with mass.
Each joint is directly torque actuated except for the point of contact with the ground, leading to one degree of underactuation for a planar model.
\ReviewRm{Next, we impose the gait hypotheses GH1-4 and GH6 in \cite{westervelt} as conditions for a walking gait.
That is, the robot must walk continuously forward in alternating phases of single support and (instantaneous) double support.
We do not require the gait to be left-right symmetric (i.e. GH5).}
The full configuration of the robot is given by the set of joint angles $\q = \{\q_1, \ldots, \q_{\nqF}\} \in \qspaceF \subset \real^{\nqF}$.
We next define the set of \emph{feasible} configurations $\qfeas \subset \qspaceF$ \ReviewEdit{(similarly to \cite{wieber2002stability})}:
\begin{definition}
    A configuration is \emph{feasible} if the joint angles satisfy actuator limits, and only foot points are touching the ground (i.e. the robot has not fallen over).
\end{definition}


Using the method of Lagrange, we can obtain a continuous dynamic model of the robot during swing phase:
\begin{equation}
\dxF(t) = \fF(\xF(t)) + \gF(\xF(t))\inptF(t).
\end{equation}
where $\xF(t) = [\q\transpose(t),\dq\transpose(t)]\transpose \in TQ \subset R^{2n_q}$ denotes the tangent space of $Q$, $u(t) \in U$, $U$ describes the permitted inputs to the system, and $t$ denotes time.

\ReviewRm{We assume that an impact event happens for each stride, following the impact hypotheses IH1-6 in \cite{westervelt}.
That is an impact is assumed to be instantaneous and impulsive, with the stance leg leaving the ground immediately after impact.}
\ReviewEdit{We assume that an instantaneous and impulsive impact occurs each time the swing foot hits the ground, with the stance leg leaving the ground immediately after impact.}
\ReviewRm{Using a similar approach to }\ReviewEdit{As in }\cite{westervelt}, we can construct a reset map for the state after impact:
\begin{align}
\xF(t^+) &:= \rxF(\xF(t^-)) \\ &= \begin{bmatrix} \rqF \q(t^-) \\ \rdqF(\q(t^-)) \dq(t^-)  \end{bmatrix}.
\end{align}
Here the superscript plus indicates the time just after the event and the superscript minus indicates the time just before the event.
$\rxF: \dqspaceF \rightarrow \dqspaceF$ is the reset map of the robot state. 
$\rqF \in \real^{\nqF \times \nqF}$ is a coordinate transformation matrix that swaps the swing leg and the stance leg after impact. 
$\rdqF: \qspaceF \rightarrow \real^{\nqF \times \nqF}$, is the configuration-dependent reset map of the configuration velocities.

This equation holds true for all states in $\sF \subset \dqspaceF$, which is called the \emph{guard} of the hybrid system, and represents the states of the robot with zero swing foot height and downwards swing foot velocity.
Any time the state of the robot enters $\sF$, the reset event must occur.

\begin{example}
The configuration $\q$ for Rabbit is shown in Figure~\ref{fig:approach} (top left).
$\qfeas$ is the set of robot configurations in which only foot points intersect the ground and all joints are within the limits: 
$q_1, q_2, q_4 \in [-\pi/2, \pi/2]$, $q_3, q_5 \in [-\pi/2, 0].$
When the swing foot intersects with the ground, we enter the guard $S$.
This causes an impulse to be transmitted to the colliding foot, and the swing and stance feet swap.
The impulse and coordinate swap are given by $\rxF$.
The joint torques torques \ReviewEdit{are saturated to} take values in the interval $\uspaceF = [-\unit[30]{Nm},\unit[30]{Nm}]^4$.
\ReviewEdit{All kinematic and inertial properties of the model are given in \cite{chevallereau2003rabbit}}.
\end{example}

\subsection{Safety}
\label{subsec:safety}
In this paper, \emph{safety} is defined as keeping the configuration \emph{feasible} for all time (i.e. $\q(t) \in \qfeas, \forall t$).
To guarantee safety, this paper finds a \emph{viability domain} \cite{wieber2002stability}:


\begin{definition}\label{def:viability}
  A \emph{viability domain} $\vdF \subset \real^{2\nqF}$ is any set satisfying $\vdF \subset \dqfeas$ which is also forward control invariant. 
  That is, there exists a Lipschitz state feedback controller $\ctrlF: \dqfeas \to \uspaceF$, such that for every initial condition $\xF_0 \in \vdF$,  the execution of the system from the initial condition remains in $\vdF$ for all time $t \in [0,\infty)$. 
  We refer to any feedback controller that is able to ensure that the system is forward control invariant as an \emph{Autonomous Viable Controller}.
\end{definition}
\noindent The forward control invariance property ensures that any state that begins within a viability domain $\vdF$ can be controlled to remain within the domain. 
Since $\vdF$ contains only feasible configurations ($\vdF \subset \dqfeas$), we know that safety can be maintained by at least one controller from all states in $\vdF$.


Once a viability domain is found, we use it to construct a semi-autonomous, safety preserving controller.
Given an initial state within $V$, a user defined control input is applied without modification to the system.
The state of the system is then continuously monitored. If the state approaches the boundary of the viability domain, the control input is overridden by an autonomous viable controller.
This gives the user full control over the system until safety is threatened, at which point, safety is automatically enforced. 
Once safety is no longer at risk, control is returned to the user.

\subsection{Goal}
\label{subsec:goal}
Using these definitions, we state our objective as:
\begin{enumerate}
    \item Find a viability domain and a corresponding autonomous viable controller.
    \item Use this domain and autonomous viable controller to construct a semi-autonomous viable controller.
\end{enumerate}


\section{Controlled Hybrid Zero Dynamics Manifold}
\label{sec:HZD}

We intend to use sums-of-squares optimization to achieve these objectives.
However, the state-space dimension of realistic robot models far exceeds the limits of this tool.
For instance, the state-space of the benchmark model Rabbit \cite{chevallereau2003rabbit} has dimension 10, while many sums-of-squares problems become computationally challenging above dimension 6 \cite{posa2017balancing}.
In this section, we show how the the state-space dimension can be reduced to a feasible size using the idea of hybrid zero dynamics \cite{westervelt}.
%

\subsection{Shaping Parameters}
The hybrid zero dynamics approach uses feedback linearization to drive the actuated degrees of freedom onto a low-dimensional manifold specified by a set of user-chosen outputs, which depend on the robot configuration $\q \in \qspaceF$.
We modify this approach by making these outputs also depend on a set of time varying shaping parameters $\aS(t) \in \aspaceS \subset \real^{\naS}$.
The shaping parameters $\aS$ are used in this paper to provide an input within the manifold dynamics.
By varying $\aS$ continuously over time, the user can change the \ReviewRm{dynamics of the} hybrid zero dynamics manifold\ReviewRm{control the robot} \ReviewEdit{to modify the robot behaviour in real-time}. 
The idea of modifying the HZD manifold \ReviewRm{parameters }in real-time is similar to \cite{ames2017first} in which the desired hip velocity $v$ acts as an input to the manifold.

We define the dynamics of $\aS$ as:
\begin{equation}
\dxS(t) = \fS(\xS(t)) + \gS(\xS(t))\inptS(t),
\end{equation}
where $\xS(t) = [\aS\transpose(t),\daS\transpose(t)]\transpose\in \daspaceS$, $\inptS(t) \in \uspaceS \subset \real^{\naS}$ are the shaping parameter inputs (with permitted values $\uspaceS$), and $t$ denotes time.
We require that $\aS$ has vector relative degree two under these dynamics.
We assume a trivial discrete update for the shaping parameters when the robot state hits a guard:  $\xS(t^+) = \xS(t^-)$.

\begin{example}
As shown in the bottom left of Figure~\ref{fig:approach}, we use a single shaping parameter $\aS(t) \in [-\pi/2,\pi/2]$ to modify the desired pitch angle of Rabbit.
\ReviewRm{This choice of shaping parameter is motivated by the observation that torso lean is an effective speed regulator for underactuated bipeds \cite{smit2017energetic}.}
\ReviewEdit{Note that this choice is somewhat arbitrary; $\aS$ could instead modify properties such as step length or center of mass height.} 
We define the dynamics of $\aS$ as follows:
\begin{equation}
\frac{d}{dt}{\begin{bmatrix} \aS(t) \\ {\daS}(t) \end{bmatrix}} = 
\begin{bmatrix}
\daS(t) \\
0
\end{bmatrix} + 
\begin{bmatrix}
0 \\ 1
\end{bmatrix}\inptS(t),
\end{equation}
where $\inptS$ represents the user-controlled pitch acceleration.
\end{example}

\ReviewEditBlock{
\subsection{Constructing the Manifold}
\label{sec:manifold}
In this subsection, we incorporate these shaping parameters in the construction of the hybrid zero dynamics manifold described in \cite{westervelt}.
Throughout the section, we use $\lfx$ and $\lgx$ to represent the Lie derivatives in $\dqspaceF$ with respect to $\fF$ and $\gF$, and $\lfa$ and $\lga$ to represent the Lie derivatives in $\daspaceS$ with respect to $\fS$ and $\gS$ (where we drop the arguments).

We begin by using a set of outputs: $h:\qspaceF\times\aspaceS\to\real^{\nuF}$ to implicitly define the hybrid zero dynamics manifold as:
\begin{multline}
    \zspaceHZD := \{(\q,\dq,\aS,\daS) \in \dqspaceF \times \daspaceS \, \rvert \, h(\q,\aS) = 0, \\
    (\lfx h)(\q,\aS,\dq) + (\lfa h)(\q,\aS,\daS) = 0\}
\end{multline}

These outputs must satisfy hypotheses similar to {HH 1-4} in \cite{westervelt}, and the resulting manifold $\zspaceHZD$ must satisfy the \emph{hybrid invariance condition}:
\begin{equation}
\label{eq:hyb_inv}
\begin{bmatrix} \rxF(\xF(t^-)) \\ \xS(t^-) \end{bmatrix} \in \zspaceHZD \quad \forall \begin{bmatrix} \xF(t^-) \\ \xS(t^-)\end{bmatrix} \in \zspaceHZD \cap \left(\sF \times \daspaceS\right).
\end{equation}

Provided these conditions are met, we can use the results in \cite[Chapter 9.3.2]{sastry2013nonlinear} to show that $\zspaceHZD$ is a smooth submanifold in $\dqspaceF \times \daspaceS$ of dimension $\nHZD = 2(\nqF - \nuF + \naS)$. 
In addition, the control input $\inptF^*:\dqspaceF \times \daspaceS \times \uspaceS \rightarrow \uspaceF$ given by:
\begin{multline}
    \label{eq:ustar}
    \hspace{-0.4cm}u^*(x,\xS,\inptS) =  -(\lgx(\lfx h + \lfa h))^{-1} \Big(\lfx(\lfx h + \lfa h) +\\
                               + \lga(\lfx h + \lfa h)\inptS + \lfa(\lfx h + \lfa h)\Big)
\end{multline}
renders $\zspaceHZD$ invariant under the hybrid dynamics of the robot (note the right hand side arguments are suppressed to simplify presentation).

As in hypothesis HH 3 in \cite{westervelt}, we define a set of phasing coordinates $\thHZD:\qspaceF \rightarrow \real^{\nqF-\nuF}$ which represent the underactuated degrees of freedom of the robot.
Using these coordinates, we can parameterize the on-manifold state of the robot  $\xHZD(t) \in \zspaceHZD$ as: $\xHZD(t) = [\thHZD(\q)\transpose, \dthHZD(\q,\dq)\transpose, \aS\transpose, \daS\transpose]\transpose$ (where we have suppressed the time dependence on the right hand side).
The continuous dynamics under this parameterization are then:
\begin{equation}
\label{eq:hatdynamics}
\dxHZD = \begin{bmatrix} \dthHZD \\ \lfx \lfx \thHZD + \lgx \lfx \thHZD \inptF^* \\ \fS + \gS \inptS \end{bmatrix} = \fHZD(\xHZD) + \gHZD(\xHZD)\inptS,
\end{equation}
where we have suppressed the time dependence.
The discrete manifold dynamics are given by:
\begin{align}
\label{eq:hzd_dyn2}
\hat{x}(t^+) &= \hat{\Delta}(\hat{x}(t^-)), \quad &&\forall \hat{x}(t^-) \in \hat{S},
\end{align}
where $t^-$ is the state before impact, and the manifold guard and reset ($\sHZD$ and $\rHZD$) are defined as:
\begin{align}
    \sHZD &= \zspaceHZD \cap \left(\sF \times \aspaceS\right)\\ 
    \rHZD(\xHZD(t^-)) &= \begin{bmatrix} \thHZD(\rqF(\q(t^-))) \\ 
    \label{eq:r_hzd}\frac{\partial \thHZD}{\partial \q}(\rqF(\q(t^-))) \rdqF(\q(t^-))\dq(t^-) \\ \xS(t^-) \end{bmatrix}.
\end{align}


\begin{example}
We begin by using the trajectory optimization toolbox FROST \cite{hereid2017frost} to find a time-varying, periodic walking trajectory: $q^{Fr}: [0,t_{max}] \to \qspaceF$.
For this trajectory, the stance leg angle of the robot: $\thHZD(\q) = -\q_1 - \q_2 - \frac{\q_3}{2}$ is monotonic in time and varies from $\thHZD_{min}$ to $\thHZD_{max}$.
This allows us to define a phasing function $t_\thHZD:[\thHZD_{min},\thHZD_{max}]\to[0,t_{max}]$ which satisfies $q^{Fr}(t_\thHZD(\thHZD(q^{Fr}(t)))) = q^{Fr}(t)$ (i.e. $t_\thHZD$ maps from points in the state space to points along the trajectory).

We modify the pitch angle of the FROST trajectory using the shaping parameter $\aS$, giving us the output function:
\begin{equation}
\targHZD(\q,\aS) = \begin{bmatrix} \q_1 - \q^{Fr}_1(t_\thHZD(\thHZD(\q))) - \aS \\[0.05cm]
                                   \q_3 - \q^{Fr}_3(t_\thHZD(\thHZD(\q))) \qquad \\[0.05cm]
                                   \q_4 - \q^{Fr}_4(t_\thHZD(\thHZD(\q))) + \aS\\[0.05cm]
                                   \q_5 - \q^{Fr}_5(t_\thHZD(\thHZD(\q))) \qquad \end{bmatrix} + h_m(\thHZD(q),\aS).
\end{equation}
Here we also added the function $h_m:\qspaceF \times \aspaceS \to \real^4$ which is chosen to ensure satisfaction of the \emph{hybrid invariance condition} \eqref{eq:hyb_inv}.
This technique for ensuring hybrid invariance is similar to the procedure given in \cite{sreenath2011compliant}.

The guard of our HZD manifold $Z$ is given as $\sHZD = \{\xHZD \mid \thHZD = \thHZD_{max}, \dthHZD > 0\}$ and the reset is defined as in \eqref{eq:r_hzd}.

\end{example}
}

\ReviewRmBlock{
\subsection{Continuous Zero Dynamics}

This subsection constructs the continuous portion of our controlled hybrid zero dynamics.
We begin by defining a set of outputs $h$, which implicitly specify points on the manifold.
Next we define a function $\thHZD$ which maps from points in the state space to points in the manifold and gives us a parametrization of our manifold states.
We then define a feedback controller $\inptF^*$ which renders the manifold invariant, allowing us to obtain continuous manifold dynamics.


Throughout the section, we use $\lfx$ and $\lgx$ to represent the Lie derivatives in $\dqspaceF$ with respect to $\fF$ and $\gF$, and $\lfa$ and $\lga$ to represent the Lie derivatives in $\daspaceS$ with respect to $\fS$ and $\gS$ (where we dropped all of the arguments).
We define the manifold implicitly as the zero level-set of a set of outputs: $h:\qspaceF\times\aspaceS\to\real^{\nuF}$, which are required to satisfy the following hypotheses (modified from \cite{westervelt}).
\begin{hhyp}
$h$ is a function of only the configuration variables and the shaping parameters;  
\end{hhyp}
\begin{hhyp}
for each point $(\q,\aS) \in \qspaceF \times \aspaceS$, the decoupling matrix $(\lgx(\lfx h + \lfa h))(\q,\aS)$ is square and invertible;
\end{hhyp}
\begin{hhyp}
there exists a smooth function $\thHZD:\qspaceF \rightarrow \real^{\nqF-\nuF}$ such that the map $\diffHZD: \qspaceF \times \aspaceS \rightarrow \real^{\nqF + \naS}$ given by $\diffHZD(\q,\aS) := [ h(\q,\aS), \thHZD(\q), \aS]\transpose$ is a diffeomorphism onto its image, and
\end{hhyp}
\begin{hhyp}
$h(\q,\aS) = 0$ for at least one point $(\q,\aS) \in \qspaceF\times\aspaceS$.
\end{hhyp}
\noindent These hypotheses allow us to construct a continuous zero dynamics manifold $\zspaceHZD$ and associated controller $\inptF^*$: 

\begin{thm}
    $\zspaceHZD := \{(\q,\aS,\dq,\daS) \in \dqspaceF \times \daspaceS \, \rvert \, h(\q,\aS) = 0, (\lfx h)(\q,\aS,\dq) + (\lfa h)(\q,\aS,\daS) = 0\}$ 
    is a smooth submanifold in $\dqspaceF \times \daspaceS$ of dimension $\nHZD = 2(\nqF - \nuF + \naS)$. The control input $\inptF^*:\dqspaceF \times \daspaceS \times \uspaceS \rightarrow \uspaceF$ given by:
    \begin{multline}
    \label{eq:ustar}
    \hspace{-0.4cm}u^*(x,\xS,\inptS) =  -(\lgx(\lfx h + \lfa h))^{-1} \Big(\lfx(\lfx h + \lfa h) +\\
                               + \lga(\lfx h + \lfa h)\inptS + \lfa(\lfx h + \lfa h)\Big)
    \end{multline}
    renders $\zspaceHZD$ invariant under the continuous dynamics (note the right hand side arguments are suppressed to simplify presentation).
\end{thm}
\noindent The proof of this theorem follows immediately from the hypotheses using results in \cite[Chapter 9.3.2]{sastry2013nonlinear}.

The diffeomorphism in HH3 implies a parameterization of the on-manifold states $\xHZD(t) \in \zspaceHZD$ as: $\xHZD(t) = [\thHZD(\q)\transpose, \dthHZD(\q,\dq)\transpose, \aS\transpose, \daS\transpose]\transpose$ (where we have suppressed the time dependence on the right hand side).
The continuous dynamics under this parameterization are then:
\begin{equation}
\label{eq:hatdynamics}
\dxHZD = \begin{bmatrix} \dthHZD \\ \lfx \lfx \thHZD + \lgx \lfx \thHZD \inptF^* \\ \fS + \gS \inptS \end{bmatrix} = \fHZD(\xHZD) + \gHZD(\xHZD)\inptS,
\end{equation}
where we have suppressed the time dependence.
It is helpful to define the coordinate transform from the manifold parameters to the ambient space $\qfHZD: Z \rightarrow \qspaceF$ as:
\begin{equation}
    \begin{bmatrix}   \qfHZD(\thHZD,\aS) \\ \aS \end{bmatrix} := \diffHZD^{-1}\begin{bmatrix} [0,\ldots,0]^T \\ \thHZD \\ \aS \end{bmatrix},
\end{equation}
i.e. $\qfHZD(\thHZD,\aS)$ is the configuration of the point $(\thHZD,\dthHZD,\aS,\daS)\in\zspaceHZD$. 


\begin{example}
We begin by using the trajectory optimization toolbox FROST \cite{hereid2017frost} to find a time-varying, periodic walking trajectory: $q^F: [0,t_{max}] \to \qspaceF$.
For this trajectory, the stance leg angle of the robot: $\thHZD(\q) = -\q_1 - \q_2 - \frac{\q_3}{2}$ is monotonic in time and varies from $\thHZD_{min}$ to $\thHZD_{max}$.
This allows us to define a phasing function $t_\thHZD:[\thHZD_{min},\thHZD_{max}]\to[0,t_{max}]$ which satisfies $q^F(t_\thHZD(\thHZD(q^F(t)))) = q^F(t)$ (i.e. $t_\thHZD$ maps from points in the state space to points along the trajectory).

We modify the pitch angle of the FROST trajectory using the shaping parameter $\aS$, giving us the output function:
\begin{equation}
\targHZD(\q,\aS) = \begin{bmatrix} \q_1 - \q^F_1(t_\thHZD(\thHZD(\q))) - \aS \\[0.05cm]
                                   \q_3 - \q^F_3(t_\thHZD(\thHZD(\q))) \qquad \\[0.05cm]
                                   \q_4 - \q^F_4(t_\thHZD(\thHZD(\q))) + \aS\\[0.05cm]
                                   \q_5 - \q^F_5(t_\thHZD(\thHZD(\q))) \qquad \end{bmatrix} + h_m(\thHZD(q),\aS).
\end{equation}
Here we also added the function $h_m:\qspaceF \times \aspaceS \to \real^4$ which ensures the satisfaction of the \emph{hybrid invariance condition} (discussed in the next section).

\end{example}

\subsection{Hybrid Zero Dynamics}

For the zero dynamics manifold to be forward invariant for the hybrid system, it must map onto itself through the guard and reset.
This gives us the \emph{hybrid invariance condition}, which states that the zero dynamics manifold $\zspaceHZD$ must satisfy:
\begin{equation}
\begin{bmatrix} \rxF(\xF(t^-)) \\ \xS(t^-) \end{bmatrix} \in \zspaceHZD
\end{equation}
for all $[\xF(t^-), \xS(t^-)]\transpose \in \zspaceHZD \cap \left(\sF \times \aspaceS\right)$.

Using the definition of Z, this can be re-stated as:
%
%
\begin{align}
    \label{eq:hyb_inv1} h(\q(t^+),\aS(t^-)) &= 0 \\
    \label{eq:hyb_inv2} \lfx h(\q(t^+),\dq(t^+),\aS(t^-)) + \hspace*{0.75cm} & \\ + \lfa h(\q(t^+), \aS(t^-), \daS(t^-)) &= 0 \nonumber
\end{align}
%
for all $[\xF(t^-), \xS(t^-)]\transpose \in \zspaceHZD \cap \left(\sF \times \aspaceS\right)$, where $[\q(t^+), \dq(t^+)]\transpose = \rxF(\xF(t^-))$.
%

If these conditions are satisfied, we call $Z$ a \emph{hybrid zero dynamics} manifold, with dynamics given by:
\begin{align}
\label{eq:hzd_dyn1}
\dot{\hat{x}}(t) &= \hat{f}(\hat{x}(t)) + \hat{g}(\hat{x}(t))\inptS, \quad &&\forall \hat{x}(t) \notin \hat{S} \\
\label{eq:hzd_dyn2}
\hat{x}(t^+) &= \hat{\Delta}(\hat{x}(t^-)), \quad &&\forall \hat{x}(t^-) \in \hat{S}.
\end{align}
$\sHZD$ and $\rHZD$ are the manifold guard and reset, respectively, defined as:
\begin{align}
    \sHZD &= \bigg\{ \xHZD\in\zspaceHZD \bigg\rvert \begin{bmatrix} \qfHZD(\xHZD) \\ \frac{d\qfHZD}{dt}(\xHZD) \end{bmatrix}\in \sF \bigg\} \\
    \rHZD(\xHZD(t^-)) &= \begin{bmatrix} \thHZD(\q(t^+)) \\ 
    \label{eq:r_hzd}\frac{\partial \thHZD}{\partial \q}(\q(t^+)) \dq(t^+) \\ \xS(t^-) \end{bmatrix}.
\end{align}
%

\begin{example}
For periodic $\q_F$, \eqref{eq:hyb_inv1} holds if $h_m(\theta_{min},\aS) = h_m(\thHZD_{max},\aS) = 0$.
We follow a similar procedure to \cite{westervelt} to ensure that \eqref{eq:hyb_inv2} is satisfied.
This procedure gives:
%
%
\begin{equation}
\label{eq:mod_targ}
h_m(\thHZD,\aS) = m(\aS)p_m(\thHZD)
\end{equation}
Where $p_m:[\thHZD_{min},\thHZD_{max}] \to \real$ is a cubic spline satisfying: 
$p_m(\thHZD_{min}) = p_m(\thHZD_{max}) = \frac{dp_m}{d\thHZD}(\thHZD_{max}) = 0$,  $\frac{dp_m}{d\thHZD}(\thHZD_{min}) = 1$
and $m: \aspaceS \to \real^4$ is given by:
\begin{equation}
m(\aS) = -\frac{\frac{\partial h}{\partial \q}^+ \rdqF(q_0^-) \frac{\partial q_0}{\partial \thHZD}^-}{\frac{\partial \thHZD}{\partial \q}^+\rdqF(q_0^-) \frac{\partial q_0}{\partial \thHZD}^-},
\end{equation}
where $\frac{\partial h}{\partial q}^+ = \frac{\partial h}{\partial q}(\thHZD_{min},\aS)$, $\frac{\partial \thHZD}{\partial q}^+ = \frac{\partial \thHZD}{\partial q}(\thHZD_{min},\aS)$, $q_0^- = q_0(\thHZD_{max},\aS)$ and $\frac{\partial q_0}{\partial \thHZD}^- = \frac{\partial q_0}{\partial \thHZD}(\thHZD_{max},\aS)$.
The guard of our HZD manifold $Z$ is given as $\sHZD = \{\xHZD \mid \thHZD = \thHZD_{max}, \dthHZD > 0\}$ and the reset is defined as in \eqref{eq:r_hzd}.

\end{example}
}
\subsection{Safety on the Manifold}
\label{subsec:safety_hzd}
We now revisit the safety criteria from Section \ref{subsec:safety} under the assumption that our state is controlled to lie on $\zspaceHZD$.
For the biped to be safe, we require that the manifold state remains in the feasible set $\qfeas$, and that the state does not leave the manifold (either by leaving the manifold boundary, or by encountering actuator limits when trying to stay on $\zspaceHZD$).
We define the \emph{unsafe states} $\fspaceHZD \subset \zspaceHZD$ as the union of:
\begin{itemize}
    \item The \emph{infeasible} states: $((\dqspaceF\setminus\dqfeas) \times \daspaceS)\cap \zspaceHZD$
    \item The states that \emph{leave the manifold boundary}, i.e. all members of the boundary set ($\partial\zspaceHZD = \{\xHZD\in\zspaceHZD \mid \qfHZD(\xHZD) \in \partial\qspaceF \text{ or } \aS \in \partial\aspaceS\}$) which do not lie on a guard, and that have an outward velocity.
    \item The states \emph{requiring unattainable actuation} to remain on $\zspaceHZD$, i.e. all states $(\xF,\xS) \in \zspaceHZD$ for which $\inptF^*(\xF,\xS,\inptS) \notin \uspaceF, \; \forall \inptS\in\uspaceS$.
\end{itemize}
Additionally we define the state-dependent set of realizable shaping parameter inputs $\uspaceHZD: \dqspaceF \times \daspaceS \to 2^{\uspaceS}$, as $\uspaceHZD(\xF,\xS) = \{\inptS\in\uspaceS\,\rvert\,\inptF^*(\xF,\xS,\inptS) \in \uspaceF\}$ (where $2^{\uspaceS}$ denotes the set of all subsets of $\uspaceS$).

Provided we constrain the manifold state to avoid $\fspaceHZD$, and constrain the shaping parameter input to lie within $\uspaceHZD$, our safety criteria is maintained.

Our goal from Section~\ref{subsec:goal} can now be re-stated as:
\begin{enumerate}
    \item Find a viability domain on $\zspaceHZD$ that does not intersect $\fspaceHZD$, and an autonomous viable controller $\ctrlHZD: \zspaceHZD\to\uspaceHZD$.
    \item Use this domain and autonomous viable controller to construct a semi-autonomous controller.
\end{enumerate}

\begin{example}
For the Rabbit example, the set of states that leave the manifold boundary are given by $\zspaceHZD_{LMB} = \{\xHZD \mid \aS = \pi/2, \; \daS > 0\} \cup \{\xHZD \mid \aS = -\pi/2, \; \daS < 0\}$. All other states on the manifold boundary either lie on a guard ($\thHZD = \thHZD_{max}, \dthHZD > 0$), or flow inwards.
We use sampling and fitting to find a region $\zspaceHZD_{Lim}\subset\zspaceHZD$ where the actuator torque limits can be satisfied for some $\inptS$.
%
We then define our unsafe set (see Fig.~\ref{fig:HZD_Set}): 
\begin{equation}
    \fspaceHZD =  (((\dqspaceF\setminus\dqfeas) \times \daspaceS)\cap \zspaceHZD) \cup \zspaceHZD_{LMB} \cup (Z \setminus \zspaceHZD_{Lim}).
\end{equation}
The set of attainable inputs $\uspaceHZD$ is given by the minimum and maximum values of $\inptS$ at each sample point $(\xF,\xS)\in\zspaceHZD $ that satisfy $\inptF^*(\xF,\xS,\inptS) \in \uspaceF$.

\end{example}

\section{Hybrid Control Invariant Set}
\label{sec:SOS}

This section outlines how the low dimensional safety problem from Section~\ref{subsec:safety_hzd} can be solved using sums-of-squares optimization \cite{parrilo2000structured}.
Broadly, the sums -of-squares approach enforces constraints of the form $p \geq 0$ (where p is a function) by constraining $p$ to be a sum-of-squares polynomial, i.e. $p = \sum_{i} p_i^2$ (where $p_i$ are polynomials). 
We refer to this constraint as $p \in SoS$.

We begin by showing how the sets and dynamics from the preceding section can be represented using polynomials.
We next define a bilinear semi-definite program for finding a viability domain, and describe the alternation used to solve it.
Finally, we construct a guaranteed safe semi-autonomous controller for the full robot, based on this viability domain.

\subsection{Polynomial Representation}
\label{subsec:sos_approx}

For the dynamics of the system to be used inside our sums-of-squares program, they must be represented in a polynomial form.
In particular, we require polynomial representations of the functions $\fHZD, \gHZD, \rHZD$ and the sets $\sHZD,\fspaceHZD, \uspaceHZD$.
Since these sets and functions can contain trigonometric as well as rational terms in their definition, we rely on approximate representations.
It is important to take care to ensure that the safety guarantee is preserved under approximation.

\ReviewEdit{
To generate polynomial approximations and verify bounding relations, we use sampling to obtain the exact function values over a dense grid in the state space.
This sampling approach is made tractable by the reduction in dimension of the previous section.
In our example, this reduces the dimension that must be sampled from 10 to 4.
We use a $30\times30\times30\times30$ sample grid to fit and bound the polynomials.
The bounds are then verified using a dense set of randomly generated test points.
}

We begin by sampling $\fHZD: \zspaceHZD \to \real^{\nHZD}$ and $\gHZD: \zspaceHZD \to \real^{\nHZD \times \naS}$ over \ReviewEdit{our} grid of points in $Z$.
Least-squares fitting can then be used to obtain the corresponding polynomial representations: $\fPoly$ and $\guPoly$.
To account for the approximation error in the continuous dynamics functions, we introduce a set of error-bounding polynomials $\gdPoly: \zspaceHZD \to \real^{\nHZD}$ which satisfy:
\begin{equation}
    \label{eq:gdpoly}
    \gdPoly(\xHZD) \geq \left\rvert \fHZD(\xHZD) -\fPoly(\xHZD) + \left(\gHZD(\xHZD) - \guPoly(\xHZD)\right)\inptHZD \right\rvert,
\end{equation}
for all $\xHZD\in\zspaceHZD$ and $\inptHZD\in\uspaceHZD$ where the inequality and absolute value are taken element-wise. 
These polynomials can be found using a linear program that minimizes the integral of $\gdPoly$ subject to \eqref{eq:gdpoly} enforced at \ReviewEdit{our} set of sample points.

To represent sets in polynomial form, we require them to take the form of semi-algebraic sets (i.e. a set $X\subset Y$ is defined as $X = \{y\in Y \mid h_i(y) \geq 0, \; \forall i=1,\ldots n\}$, where $h: Y \to \real^{n}$ is a collection of polynomials).  
We use a bounding set to approximate the reset map $\rHZD: \sHZD \to \zspaceHZD$ in a conservative manner.
That is, we find a set $\rPoly \subset \zspaceHZD \times \zspaceHZD$ that bounds all possible reset behaviours: 
\begin{equation}
(\xHZD,\rHZD(\xHZD)) \in\rPoly,\; \forall \xHZD\in\sHZD.
\end{equation}
The sets $\sHZD$ and $\fspaceHZD$ are represented with semi-algebraic outer approximations as follows: $\fspaceHZD_p \supset \fspaceHZD$, $\sHZD_p \supset \sHZD$.
We define the sets $ \rPoly, \fspaceHZD_p, \sHZD_p$ using the respective polynomials: $h_R: \zspaceHZD \times \zspaceHZD \to \real^{n_{hr}}$,  $h_F: \zspaceHZD \to \real^{n_{hf}}$, $h_S: \zspaceHZD \to \real^{n_{hs}}$. 
The space of feasible inputs $\uspaceHZD$ can be approximated using a state-dependent box constraint: \begin{equation}
\inptHZD_{min}(\xHZD) \leq \inptHZD(\xHZD) \leq \inptHZD_{max}(\xHZD), \; \forall \xHZD\in\zspaceHZD \setminus \fspaceHZD_p
\end{equation}
where $\inptHZD_{min},\inptHZD_{max}: \zspaceHZD \setminus \fspaceHZD_p \to \uspaceS$ are polynomial input bounds, and the inequality is taken element-wise.
The set of inputs that satisfy this box constraint is denoted by $\uspaceHZD_p$.

\subsection{Optimization Formulation}

We use an optimization similar to \cite{posa2017balancing} to find the largest possible viability domain $\vdHZD \subset \zspaceHZD \setminus \fspaceHZD$ for our hybrid zero dynamics system.
We represent $\vdHZD$ as the zero super-levelset of a polynomial function $\bfHZD:\zspaceHZD \to \real$ (i.e. $\vdHZD = \{\xHZD\in\zspaceHZD \mid \bfHZD(\xHZD) \geq 0\}$), and represent the autonomous viable controller using a polynomial function $\ctrlHZD:\zspaceHZD \to \real^{\naS}$.
To enforce the viability of $\vdHZD$ according to Definition~\ref{def:viability}, we require $\bfHZD$ and $\ctrlHZD$ to satisfy four conditions:

\begin{viability} \text{ }
\begin{enumerate}
    \item $\vdHZD$ does not intersect $\fspaceHZD$ (i.e. $\bfHZD(\xHZD) < 0, \; \forall \xHZD\in \fspaceHZD$)
    \item All states that are contained in both the guard and $\vdHZD$ must be mapped to a state in $\vdHZD$ (i.e. $\bfHZD(\rHZD(\xHZD)) \geq 0, \; \forall \xHZD \in \{\xHZD\in \sHZD \mid \bfHZD(\xHZD) \geq 0\}$)
    \item At the boundary of $\vdHZD$ (i.e. where $\bfHZD(\xHZD) = 0$), the state flows inward under the controller $\ctrlHZD$ (i.e. $\frac{d\bfHZD}{dt} > 0$)
    \item The autonomous safe controller must satisfy the input bounds within the safe set (i.e. $\ctrlHZD(\xHZD)\in\uspaceHZD,\; \forall \xHZD \in \vdHZD$)
\end{enumerate}
\end{viability}

Condition 1 ensures that states can not leave the viability domain by simply leaving the space $\zspaceHZD$. 
Condition 2 ensures that states can not leave the viability domain when traversing a guard.
Condition 3 ensures that states cannot leave the viability domain under the continuous dynamics of the system.
Finally, Condition 4 ensures that our controller respects the robot torque constraints.
Each of these conditions are ensured with a corresponding sums-of-squares constraint, giving us: 

\begin{constraint} (Viability Condition 1)
$$-\bfHZD -\sPoly{1} h_F \in \sos$$
\end{constraint}
Here $\sPoly{1}:\zspaceHZD\to\real^{1\times n_{hf}} \in \sos$ are \ReviewEdit{sums-of-squares polynomials} \ReviewRm{s-functions \cite{parrilo2000structured}} that relax the positivity constraint outside $\fspaceHZD_p$.
\ReviewEdit{We refer to such polynomials as \emph{s-functions}.} 

\begin{constraint} (Viability Condition 2)
$$\bfHZD^+ - \bfHZD^- - \sPoly{2} h_R - \sPoly{3} h_S^- \in \sos$$
\end{constraint}
Here $\sPoly{2}:\zspaceHZD\times\zspaceHZD\to\real^{1\times n_{hr}},\sPoly{3}:\zspaceHZD\to\real^{1\times n_{hs}} \in \sos$ are s-functions.
The superscripts $-$ and $+$ indicate whether a function is evaluated using the first ($-$) or second ($+$) argument of $h_R:\zspaceHZD\times\zspaceHZD\to\real^{n_{hr}}$.
That is, this constraint enforces: $\bfHZD(\xHZD^+) - \bfHZD(\xHZD^-) - h_R(\xHZD^-,\xHZD^+)\sPoly{2}(\xHZD^-,\xHZD^+) - h_S(\xHZD^-)\sPoly{3}(\xHZD^-) > 0,\; \forall (\xHZD^-,\xHZD^+) \in \zspaceHZD\times\zspaceHZD$.
\ReviewEdit{Note that the addition of the $\sPoly{3}$ term is not strictly necessary, since points in $\sHZD_p$ must lie in $\rPoly$.
However, this term can help relax the constraint when points in $\rPoly$ lie outside $\sHZD_p$.}

\begin{constraint} (Viability Condition 3)
\begin{align*}
    \lfPoly \bfHZD + \lguPoly \bfHZD \ctrlHZD + \sum_{j=1}^{\nHZD} \qPoly_{j} + \bfHZD \lPoly + \sPoly{4} h_F &\in \sos\\
    \qPoly - \lgdPoly \bfHZD + \sPoly{5} h_F &\in \sos\\
    \qPoly + \lgdPoly \bfHZD + \sPoly{6} h_F &\in \sos
\end{align*}
\end{constraint}
Here $\sPoly{4}:\zspaceHZD\to\real^{1\times n_{hf}} \in \sos$ and $\sPoly{5},\sPoly{6}:\zspaceHZD\to\real^{n_d \times n_{hf}}$ are s-functions that relax the constraint inside $\fspaceHZD_p$, and $\lPoly:\zspaceHZD\to \real$ is a slack polynomial that can relax the constraint whenever $\bfHZD \neq 0$.
The polynomials $\qPoly:\zspaceHZD\to\real^{\nHZD}$ are used to bound the effects of the dynamics error $\gdPoly$ on the time derivative of $\bfHZD$.

\begin{constraint} (Viability Condition 4)
\begin{align*}
    \ctrlHZD - \inptHZD_{min} + h_F\sPoly{7} \in \sos \\
    -\ctrlHZD + \inptHZD_{max} + h_F\sPoly{8} \in \sos
\end{align*}
\end{constraint}
Here $\sPoly{7},\sPoly{8}:\zspaceHZD\to\real^{\naS} \in \sos$ are s-functions that relax the constraint inside $\fspaceHZD_p$.

The desired objective of our optimization is to maximize the volume of $\vdHZD$.
This volume is difficult to compute exactly for an arbitrary $\bfHZD$, since the domain of integration is given by a semi-algebraic set.
We propose an analytically tractable approximation to this objective:
\begin{equation}
\int_{\zspaceHZD} \label{eq:objective} \bfHZD(\xHZD) d\xHZD.
\end{equation}
This objective is combined with the following constraint in order to approximate the volume of $\vdHZD$:
\begin{constraint} (Objective Constraint)
\label{eq:const_obj} $$1 - \bfHZD \in \sos.$$
\end{constraint}
To understand how this objective and constraint approximate the volume of $\vdHZD$, take a continuous function $\bfHZD$ that satisfies the constraints of the previous section.
%
For every point $\xHZD$ not in the set $\fspaceHZD$, the value $\bfHZD(\xHZD)$ is constrained only by Constraint \eqref{eq:const_obj}.
This means that  $\bfHZD(\xHZD)$ can increase to a value of 1 for points inside $\vdHZD$, and $\bfHZD(\xHZD)$ increases to a value of 0 for points outside this set.
As a result, $\bfHZD$ approaches the indicator function over $\vdHZD$, and the integral in the objective function approaches the volume of $\vdHZD$.


Combining the constraints and objective, we arrive at the following sums-of-squares problem:
\begin{align}
    &\underset{\begin{subarray}{c}
    \bfHZD,\ctrlHZD,\qPoly,\lPoly\\
    \sPoly{1},\ldots,\sPoly{6} \end{subarray}}{\sup} \; \int_{\zspaceHZD} \bfHZD(\xHZD) d\xHZD 
    \label{eq:bl_SOS} \\ 
    &\begin{aligned}
        \mathrm{s.t.}\; 
                     &\mathrm{SoS\;Constraints}\; 1 {-}5, \\ &\sPoly{1},\ldots,\sPoly{8} \in SoS \nonumber
    \end{aligned}
\end{align}
To express this problem as a semi-definite program or SDP (which can be solved with commercial solvers), all $\sos$ constraints must be linear functions of the decision variable polynomials.
However, Constraint 3 in the above problem includes the terms $\lguPoly \bfHZD \ctrlHZD$ 
and $\lPoly \bfHZD$ which are \emph{bilinear} in $\ctrlHZD,\,\bfHZD$ and in $\lPoly,\,\bfHZD$ respectively.
Problems of this form are referred to as bilinear sums-of-squares problems.
The bilinear nature of the constraints means that these problems are non-convex, and we can no longer guarantee a globally optimal solution to this problem.

To solve this nonconvex bilinear sums-of-squares program we turn to a strategy called alternation. 
This strategy breaks \eqref{eq:bl_SOS} into a pair of linear sums-of-squares programs which can each be solved using a commercial solver.
In each program one of the bilinear variables is kept fixed while the other is optimized over. 
The variables that were optimized are then fixed while the other pair of variables are optimized.
If the final solution satisfies the constraints of the original program, the solution is guaranteed to be a viability domain. 
Computationally, each SDP is formulated in spotless\footnote{\url{https://github.com/spot-toolbox/spotless}} and solved using Mosek.

\subsection{Guaranteed Safe Semi-autonomous Controller}
\label{sec:controller}

We use a feasible solution to the above optimization problem to generate a guaranteed safe semi-autonomous controller.
This controller modifies user input to ensure that the Viability Conditions 3 and 4 are always satisfied.
Condition 4 can be enforced by saturating the user inputs to always lie within the input bounds. 
To enforce condition 3, we note that it is only active on the boundary of $\vdHZD$.
This means that we can ensure safety so long as we use the autonomous safe controller $\ctrlHZD$ when the state lies on the boundary of $\vdHZD$, i.e. $\{\xHZD\in\zspaceHZD \rvert \bfHZD(\xHZD) = 0\}$.

Since a controller that is discontinuous on the boundary of the safe set would pose difficulties for systems with finite bandwidth, we additionally must ensure that the new controller is continuous near the boundary.
To achieve this, we smoothly interpolate between the user input $\rawctrlPoly$ and the guaranteed safe controller $\ctrlHZD$ (which satisfies the safety condition when $\bfHZD(\xHZD) = 0$) to get the regulated input $\maskctrlPoly$:
\begin{equation}
    \maskctrlPoly = \rawctrlPoly + (\ctrlHZD(\xHZD) - \rawctrlPoly)\wsafePoly(\bfHZD(\xHZD),\epsilon),\label{eq:safe}
\end{equation}
where $\ctrlHZD$ and $\bfHZD$ are computed using \eqref{eq:bl_SOS},
$\wsafePoly:\real\to [0,1]$ is a smooth step-like function that satisfies $\wsafePoly(v,\epsilon) = 0,\,\forall v \geq \epsilon$, and $\wsafePoly(v,\epsilon) = 0,\,\forall v \leq \epsilon/2$, and $\epsilon \in (0,1)$ controls the smoothness of the interpolation.

When $\xHZD$ satisfies $\bfHZD(\xHZD) > \epsilon$, the user input is unmodified, as we are sufficiently removed from the boundary of the safe set. When $0 \leq \bfHZD(\xHZD) \leq \epsilon/2$, the safe controller is fully active, keeping the state in the safe set.

%% file: sections/Results.tex
\section{Results}
\label{sec:results}

\begin{figure}
    \centering
    \includegraphics[width=0.95\columnwidth]{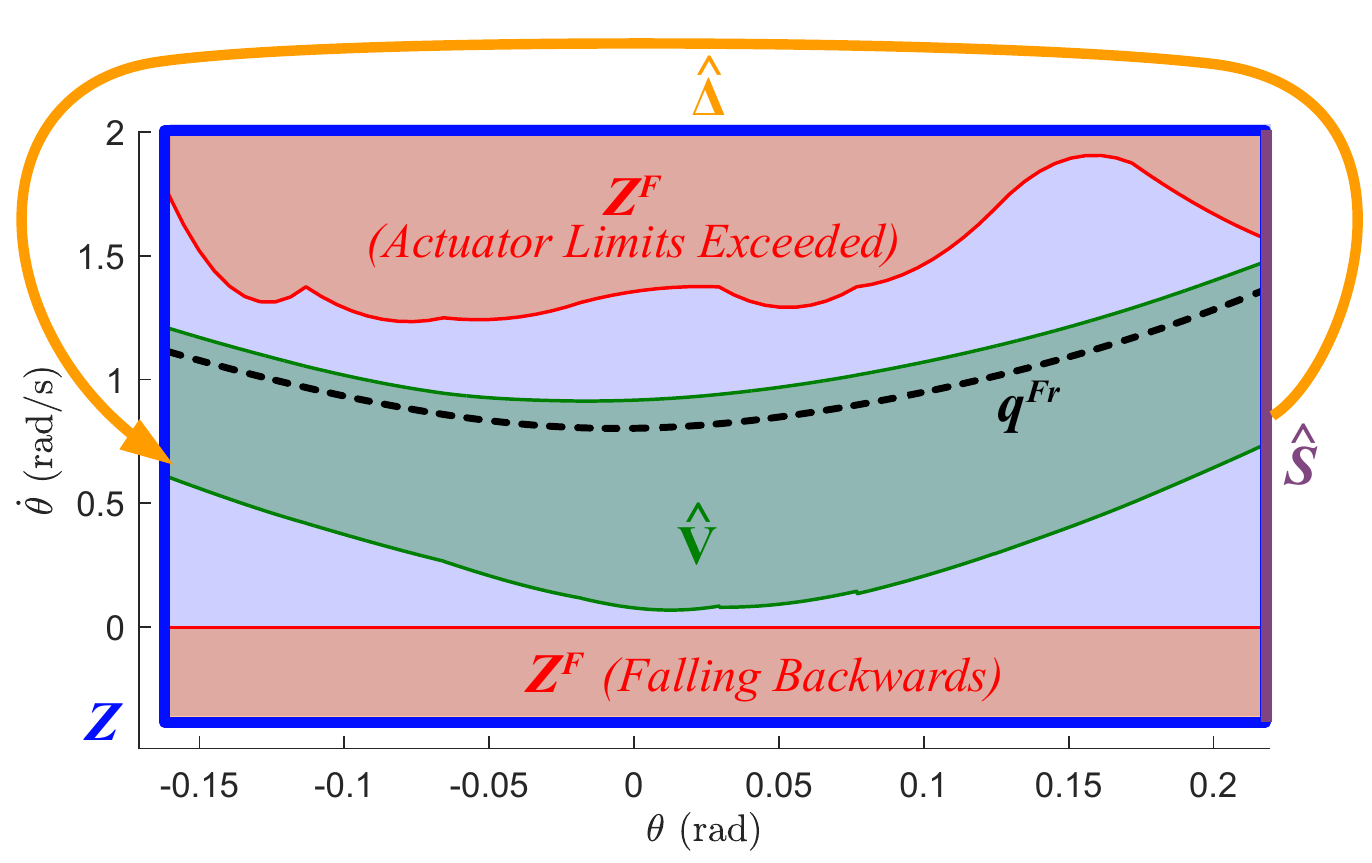}
    \vspace*{-0.4cm}
    \caption{\ReviewEditRm{(Figure Updated: $q^{Fr}$ added) } A 2D slice (along $\aS = \daS = 0$) of the four-dimensional viability domain $\vdHZD$ (shown in green) for Rabbit. 
    \ReviewRm{The state flows from left to right in the figure.}
    The border at the right corresponds to the hybrid guard $\sHZD$ of foot touchdown, where the state is reset under the map $\rHZD$ to the left of the figure.
    The unsafe set $\fspaceHZD$ is shown in red.
    \ReviewEdit{We avoid the lower region ($\dthHZD < 0$) in order to conservatively prevent backwards falls.}
    The upper region conservatively approximates the region in which the control input \eqref{eq:ustar} violates the torque limits of the robot.
    \ReviewEdit{By modifying the control input whenever Rabbit is at the edge of $\vdHZD$, $\fspaceHZD$ can be avoided indefinitely.
    This is illustrated in the attached video.}
    \ReviewEdit{Finally, the periodic trajectory used to generate our targets $q^{Fr}$ is shown in dashed black.
    Note that our viability domain is able to guarantee robot safety even for states far away from this nominal trajectory.}}
    \label{fig:HZD_Set}
    \vspace*{-0.6cm}
\end{figure}

We used the proposed approach to compute a viability domain for the robot Rabbit \cite{chevallereau2003rabbit}.
The viability domain is represented using a set of 8 degree-4 polynomials, each covering an interval within the full range of $\thHZD$.
A two-dimensional slice of the viability domain $\vdHZD$ is shown in Fig.~\ref{fig:HZD_Set}.

To demonstrate the semi-autonomous safe controller, we used it to ensure safety while performing a reference following task.
The task is to track a time-varying pitch angle $\aS_d:[0,\infty)\to\aspaceS$.
To follow the target, we set a desired pitch acceleration $\inptS$ using a "na\"ive" PD controller:
\begin{equation}
    \label{eq:naive}
    \inptS^{d}(\xS,t) = k_p (\aS_{d}(t) - \aS) + k_d (\dot{\aS_{d}}(t) - \daS) + \ddot{\aS_{d}}(t).
\end{equation}

We used the feedback controller \eqref{eq:ustar} to map the desired acceleration to the four motor torques of the Rabbit model.

For the feedback controller to respect Rabbit's actuator torque limits, we first saturated $\inptS$ with a real-time Quadratic Program (QP) to get the input to our safety regulator:
\begin{align}
    \rawctrlPoly(\xF,\xS,t) = &\underset{\inptS}{\min} \; \left\rvert \inptS - \inptS^d(\xS,t) \right\rvert^2\label{eq:sat}\\ 
    &\mathrm{s.t.}\; u^*(\xF, \xS, \inptS) \in [-\unit[30]{Nm},\unit[30]{Nm}]^4 \nonumber
\end{align}

Using a QP to satisfy the actuator constraints of the system is similar to many state of the art approaches for high-dimensional robot control \cite{hsu2015control,nguyen2015optimal,nguyen2016exponential,nguyen2016dynamic} .
A major limitation of these approaches is the inability to guarantee the feasibility of the QP.
That is, for some states, there may not be an input that satisfies the actuator constraints (the set of such states is shown in red in Fig.~\ref{fig:HZD_Set}).

\begin{figure}
    \centering
    \includegraphics[width=\columnwidth]{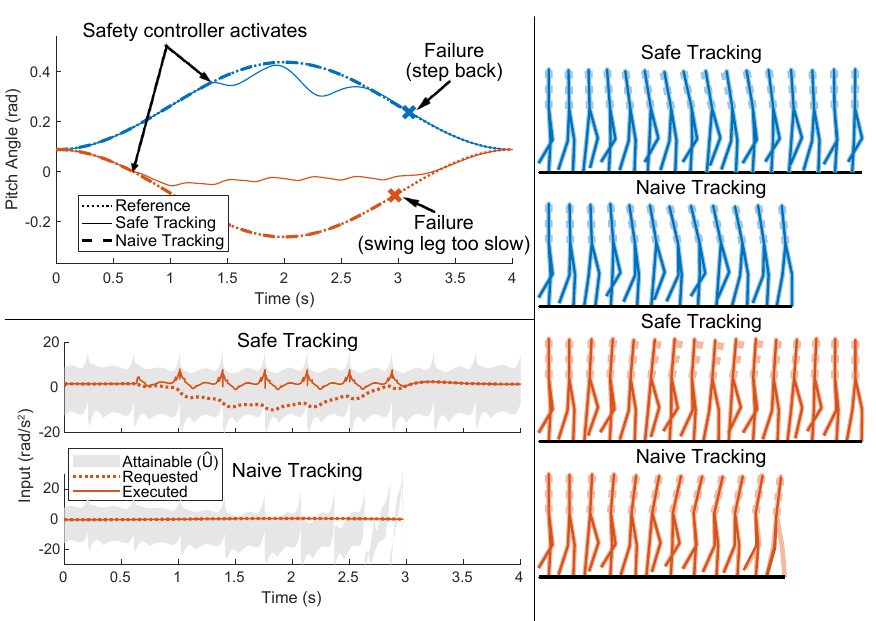}
    \vspace*{-0.5cm}
    \caption{\ReviewEditRm{(Figure Updated: legend fixed)} Tracking performance of the safe \eqref{eq:safe} and na\"ive \eqref{eq:naive} controllers following two reference trajectories under the full rabbit dynamics. 
    The pitch angles are shown in the top left. 
    For both references, the safe controller modifies the input before safety is at risk, while the na\"ive controller follows the reference even as it leads to failure.
    Failure for the upper trajectory corresponds to stepping backwards, and in the lower trajectory corresponds to moving too fast for the swing leg to reach its target. 
    The bottom left figure shows desired input $\inptS^d$ and executed input for both nai\"ive and safe tracking controllers following the second reference target.
    The state-dependent region of inputs that satisfy the torque constraints \ReviewRm{under feedback linearization }are shown in grey. 
    Note that under the na\"ive controller, this region vanishes as the forward walking speed of the robot becomes too high.
    Stills from the simulation trajectories are shown on the right.
    The dotted line is the desired pitch, and the faded line is the nearest on-manifold state $q_0(\thHZD,\aS)$.
    \ReviewEdit{See the attached video for an animated presentation of these results.}
    }
    \label{fig:tracking}
    \vspace*{-0.65cm}
\end{figure}

Our approach guarantees the feasibility of \eqref{eq:sat} by constraining the state of the robot to be within the QP-feasible region (i.e. outside of $\fspaceHZD$ in Fig.~\ref{fig:HZD_Set}).
To maintain this state constraint, we modified the input $\inptHZD_0$ using the guaranteed safe semi-autonomous controller defined in \eqref{eq:safe}.
In Fig.~\ref{fig:tracking}, we compare the results of the na\"ive controller \eqref{eq:naive} and the safe controller \eqref{eq:safe} using a simulation of the full dynamics of the robot Rabbit.
\ReviewRm{Two desired pitch angle trajectories are shown, with the robot temporarily pitching either backwards or forwards.}

When tracking the backwards pitch target, the na\"ive controller slows to the point of falling backwards, while the safe controller deviates slightly to maintaint forward walking.
For the forward pitch target, the na\"ive controller speeds up as it leans forward.
At a certain speed, it cannot longer stay on the low dimensional manifold under the torque limits and falls.
The safe controller recognizes this risk early and deviates from the desired forward pitch before reaching this speed.
The bottom-left figure shows how the set of torque-limit satisfying control inputs disappears for the na\"ive controller.

\ReviewEdit{
This task demonstrates that robot safety can be maintained even for states that are far away from any periodic limit cycle.
Indeed, the only periodic limit cycle used in our approach keeps the body pitch relatively upright ($\aS = 0$).
As such, our approach broadens the set of real-time safe behaviours that can be executed by Rabbit, since previous methods \cite{ames2014rapidly,nguyen2015optimal,nguyen2016dynamic,motahar2016composing} would all require a pre-computed limit cycle for each new reference trajectory.
}

%% file: sections/Discussion.tex
\section{Conclusion}
\label{sec:conclusion}

This paper presents a method to construct a guaranteed safe semi-autonomous controller for high-dimensional walking robots.
The resulting controller guarantees viability and allows for flexible input when viability is not at risk.
The method is evaluated on a model of the robot Rabbit, and a tracking task is used to illustrate its capabilities.
With a 10-dimensional state space, this model is larger than any known model for which continuous-time safety guarantees have been generated.

\ParHeader{Extensions to complex systems}
\ReviewEdit{
Despite this increase in model dimension, our example is still somewhat simplified: the dynamics are two dimensional, the terrain is flat, and the range of behaviour is limited to modifying the torso pitch angle.
In contrast, bipedal robots in the world must traverse three dimensional, varied terrain while performing a wide range of tasks.
}

\ReviewEdit{
When extending our method to these cases, a trade-off arises between the degree of underactuation of the model, the genericity of the behaviour (i.e. the number of shaping parameters), and the computational complexity of the optimization problem.
From Section \ref{sec:manifold}, the dimension of the reduced order manifold (our state space) is twice the sum of the degree of underaction and the number of shaping parameters.
In \cite{posa2017balancing}, the authors show that a 6 dimensional state space is tractable for similar sums-of-squares programs.
Our approach can thus currently handle a maximum of three degrees of underactuation and/or shaping parameters.
}

\ReviewEdit{
Under this constraint, we can directly extend our method to 3D. 
For instance, take the 3D biped with controlled steering given in \cite{motahar2016composing}.
This application has two degrees of underactuation (pitch and yaw) and would have one shaping parameter controlling yaw rate (i.e. turning left or right).
Using our method, we could construct a safe steering controller for the robot that avoids the risk of turning too quickly and falling. 
An extension to rough terrain, however, will likely require improvements in scaling of the sums-of-squares problem.
Such scaling improvements are an active research target \cite{papp2019sum,ahmadi2014dsos}.
}

The core insight behind our approach is that sums-of-squares and hybrid zero dynamics are remarkably complementary tools.
Sums-of-squares analysis generates the set based guarantees needed to render hybrid zero dynamics safe, and hybrid zero dynamics provides the dimensionality reduction needed for sums-of-squares analysis to be tractable.
The key innovation for combining these two tools was the introduction of a set of shaping parameters which control the dynamics on the manifold. 
The ability to combine sums-of-squares and hybrid zero dynamics presents a promising path forward for building guaranteed safe walking controllers for complex legged robots.

